\documentclass{article}

\usepackage{amssymb}
\usepackage{graphicx}
\usepackage{multirow}
\usepackage{amsmath}
\usepackage[font=small,labelfont=bf]{caption}

\usepackage{url}
\urldef{\mailsa}\path|lisa.graziani@unifi.it|
\urldef{\mailsb}\path|{mela,marco}@diism.unisi.it|

\begin{document}
\title{Coherence Constraints in\\ Facial Expression Recognition} 
\date{}

\author{ 
	Lisa Graziani \\
	\small DINFO, University of Florence, Italy\\
	\small \texttt{lisa.graziani@unifi.it} \\
	\and
	Stefano Melacci \\
	\small DIISM, University of Siena, Italy \\
	\small \texttt{mela@diism.unisi.it} \\
	\and
	Marco Gori \\
	\small DIISM, University of Siena, Italy \\
	\small \texttt{marco@diism.unisi.it}
}

\maketitle

\begin{abstract}
Recognizing facial expressions from static images or video sequences is a widely studied but still challenging problem. The recent progresses obtained by deep neural architectures, or by ensembles of heterogeneous models, have shown that integrating multiple input representations leads to state-of-the-art results. In particular, the appearance and the shape of the input face, or the representations of some face parts, are commonly used to boost the quality of the recognizer. 
This paper investigates the application of Convolutional Neural Networks (CNNs) with the aim of building a versatile recognizer of expressions in static images that can be further applied to video sequences. We first study the importance of different face parts in the recognition task, focussing on appearance and shape-related features. Then we cast the learning problem in the Semi-Supervised setting, exploiting video data, where only a few frames are supervised. The unsupervised portion of the training data is used to enforce three types of coherence, namely temporal coherence, coherence among the predictions on the face parts and coherence between appearance and shape-based representation. Our experimental analysis shows that coherence constraints can improve the quality of the expression recognizer, thus offering a suitable basis to profitably exploit unsupervised video sequences.
Finally we present some examples with occlusions where the shape-based predictor performs better than the appearance one. 
\end{abstract}

\section{Introduction}
\label{sec:intro}

Facial expression recognition is the problem of detecting emotions in facial images or videos. The research activity on this problem involves the scientific community that is about psychology but also the one that is about computer science and artificial intelligence. Although this task is widely studied and much progress has been made, it still remains a challenging problem, due to the variability and complexity of facial expressions. As a matter of fact, facial expressions can be categorized with respect to multiple classes of emotions. The most widely followed approach consists in considering six basic emotions plus the neutral case, and it is due to the studies of Paul Ekman \cite{ekman1971constants}, while other scientists provided more fine grained descriptions \cite{plutchik2001nature}.
Facial features of expressions are mostly located around mouth, nose, and eyes, and their locations are essential in explaining and categorizing expressions \cite{duchenne1990mechanism}. Despite the large number of advanced psychological experiments about the human perception and recognition of emotions, we can trivially figure out that different face parts have a different impact in the way humans recognize emotions: the role of eyebrows when we are angry or the way we treat our mouth when we are happy or surprised, for example. 

We can find several approaches that exploit Machine Learning with the aim of learning to categorize emotions from examples. Most of them are about using still images \cite{lopes2017facial,mollahosseini2016going}, while several more recent works also consider video sequences where actors start with a neutral expression and generate a non-neutral one \cite{long2016video,zhang2017facial}. The learning framework is usually fully supervised, and supervision is either about each training image or about each video sequence. Works that exploit video data focus on the importance of the temporal evolution of the input face.
The system proposed by Fan and Tjahjadi \cite{fan2015spatial} processes four sub-regions of the face: forehead, eyes/eyebrows, nose and mouth.
They used an extension of the spatial pyramid histogram of gradients and dense optical flow to extract spatial and dynamic features from video sequences, and adopted a multi-class SVM-based classifier with one-to-one strategy to recognise facial expressions.
Jung et al. \cite{jung2015joint} propose a neural-network-based method where two different networks are exploited: the first one extracts appearance features from image sequences, learning temporal correlations, while the other network extracts shape features from a set of facial landmarks. The two nets are combined to yield the final decision on the emotion class. 
Happy and Routray \cite{happy2015automatic} identify salient areas with generalized discriminative features for expression classification. They only use appearance-based features, and they do not consider the time domain. 
The framework from Jain et al. \cite{jain2011facial} recognizes facial expressions from video sequences by modeling temporal variations within shapes. They show that shape provides important information that is sometimes hard to grasp from appearance only.
Zhang and Huang \cite{zhang2017facial} propose a mixed model which include a ``temporal'' and a ``spatial'' network. The former captures dynamic features from consecutive frames, while the latter is about extracting static features from still frames.
More generally, we can roughly characterize the popular trends in the existing literature by the usage of (\textit{i.}) appearance-related (i.e., visual) features, (\textit{ii.}) shape-related features, (\textit{iii.}) features from face parts, (\textit{iv.}) the temporal domain (i.e., video data).

This paper investigates the application of a pool of Convolutional Neural Networks (CNNs) with the aim of building recognizers of expressions in static images, that can be further applied to video sequences. We consider both (\textit{i.}) appearance and (\textit{ii.}) shape features, but, differently from most of the existing works, we do not hand-engineer shape features, and we let the CNNs learn the right representations from special shape-only images. 
We show that shape-based representation can help the expression recognition when there are some occlusions on the face. 
We propose a model that considers (\textit{iii.}) sub-parts of the face in addition to the entire face, motivated by the need of gaining deeper insights in the role of each component. Then, we move to the Semi-Supervised setting, exploiting (\textit{iv.}) video data. The unsupervised portion of the training data is used to enforce ``temporal coherence'' among consecutive frames, ``part coherence'' in each frame, i.e., a coherent prediction among the CNNs that operate on the different face parts, and coherence between appearance and shape-based representation for each face part. 
Our experimental analysis shows that coherence constraints can improve the quality of the expression recognizer, thus offering a suitable basis to profitably exploit unsupervised video sequences.

Finally we present some examples to show that the shape-based representation can help to detect the right expression when the appearance-based representation fails, such as in presence of occlusions in some face parts as mouth or nose. 

This paper is organized as follows. The next Section formalizes the problem of facial expression recognition. Section \ref{sec:model} introduces our model. The role of coherence is described in Section \ref{sec:coherence}, while experiments are collected in Section \ref{sec:experiments}. Section \ref{sec:occlusions} reports some experiments on frames with occlusions and finally in Section \ref{sec:conclusions} are reported the conclusions (and future work).

\section{Facial Expression Recognition}
\label{sec:rec}


The task of facial expression recognition that we consider in this paper consists in building a classifier that predicts one of the six universal emotions \cite{ekman1971constants}, that are \textit{anger, disgust, fear, happiness, sadness, surprise}, plus the \textit{neutral} case, and that we collect into the set $Y$, codified with indices from $1$ to $7$.
The most popular inputs of the recognizer are images of faces, represented in foreground, usually with frontal orientation.
When video data is considered, the recognition problem focusses on short video clips where a transition from the \textit{neutral} state toward one the six emotions is recorded.
Processing videos instead of still images can improve the recognition performance because facial expressions involve variations of the facial muscles along the temporal dimension. However, classifiers that are specifically trained to build a latent representation from a video clip $\mathcal{V}$ before taking a decision \cite{jung2015joint}, cannot be immediately applied to classify images. Differently, image-based classifiers can process single frames $\{ \mathcal{I}_t \}$ of a video (being $t$ the time index) to produce a final decision over a time window, so they are more versatile from the point of view of easiness of deployment in different real-world applications. 
The facial expression recognition problem is usually faced in the ``Fully-Supervised'' setting, and, in the case of videos, the available datasets are composed of labeled video clips where we do not have access to the labelings of the single frames\footnote{See CK+ \url{http://www.consortium.ri.cmu.edu/ckagree/
}, Oulu-CASIA \url{http://www.cse.oulu.fi/CMV/Downloads/Oulu-CASIA}, MMI \url{https://mmifacedb.eu/}.}. 
Nonetheless, obtaining supervised data is costly, while nowadays is pretty easy to have access to collections of unsupervised frontal view faces (web, social networks, smartphones, ...) or unsupervised video recordings (video conference/call applications). This suggests that studying the ``Semi-Supervised'' setting, where a portion of the training data is labeled and a larger portion is unsupervised, can be a promising way to approach the recognition task. 

Motivated by the need of building a versatile emotion recognition system, we focus on a predictor that operates on still images and that we can use to make predictions on video data. The system can be trained exploiting both video and image data in a Semi-Supervised setting, taking advantage of the temporal evolution described by the video format.
In detail, we consider a classifier $f(\cdot)$ that produces a decision $y \in Y$ for each input image $\mathcal{I}$, or for a set of consecutive frames belonging to a time window $W$ (that covers a video clip, for example),
\begin{eqnarray}\label{classifier}
\label{a}y &=& f(\mathcal{I})\\
\label{b}y &=& \texttt{majority}_{t\in W} \left\{ f(\mathcal{I}_t) \right\} \ ,
\label{f}
\end{eqnarray}
where \texttt{majority} is the majority-voting function, that returns the most frequent prediction in the time window $W$.
Differently from the existing approaches, our system can be trained using labeled and unlabeled image databases, collected in $\mathcal{D}_{\mathcal{I}}$, or labeled and unlabeled frames extracted from the previously described labeled video sequences, collected in $\mathcal{D}_{\mathcal{V}}$. Due to the aforementioned properties of the existing video datasets (containing transitions from \textit{neutral} to a certain emotion), we can artificially generate $\mathcal{D}_{\mathcal{V}}$ by labeling as \textit{neutral} the very first frames of each video clip, and by assigning the provided video label to the last frames of the sequence. The frames in the internal portion of the sequence are not labeled. Formally, we have
\[
\mathcal{D}_{\mathcal{I}} = \{ (\mathcal{I}_i,y_i),\ i=1,\ldots,l \}\ \cup\ \{ (\mathcal{I}_i, \texttt{none}),\ i=l+1,\ldots,l+u\} \ ,
\]
where $y_i \in Y$ is the image label, and the rightmost set is fully unlabeled. Then,
\[
\mathcal{D}_{\mathcal{V}} = \{ \mathcal{D}_{\mathcal{V}_z},\ z=1,\ldots,v \} \ ,
\]
where $v$ is the number of available video clips and $\mathcal{D}_{\mathcal{V}_z}$ is a sequence extracted from the $z$-th clip,
\begin{eqnarray*}
\mathcal{D}_{\mathcal{V}_z} &=& \left( (\mathcal{I}_{z,t},\texttt{neutral}),\ t=1,\ldots, \alpha|\mathcal{V}_z| \right) \oplus  \\ 
& & \left( (\mathcal{I}_{z,t}, \texttt{none}),\ t=\alpha|\mathcal{V}_z|+1,\ldots,\beta|\mathcal{V}_z| \right) \oplus  \\
& &  \left( (\mathcal{I}_{z,t},y_z),\ t=\beta|\mathcal{V}_z|+1,\ldots,|\mathcal{V}_z| \right) \ ,
\end{eqnarray*} 
being $\oplus$ the sequence concatenation operator, $\mathcal{I}_{z,t} \in \mathcal{V}_z$ the $t$-th frame of the $z$-th video, and $0 < \alpha < \beta < 1$, arbitrarily chosen. In this case $y_z \in Y \setminus \{\texttt{neutral}\}$ is the label provided with the video clip $\mathcal{V}_z$ (\texttt{neutral} is the identifier of the \textit{neutral} class).
We notice that $\mathcal{D}_{\mathcal{V}}$ is more informed than $\mathcal{D}_{\mathcal{I}}$, since it also stores the image/frame order and the frame grouping with respect to the videos. For this reason, we can consider $\mathcal{D}_{\mathcal{I}}$ to be an instance of the more general representation $\mathcal{D}_{\mathcal{V}}$, and in the rest of the paper we will focus on data represented as in $\mathcal{D}_{\mathcal{V}}$ without reducing the generality of what we described so far, and we will compactly indicate it with $\mathcal{D}$.

\section{Model}
\label{sec:model}

Our model is based on CNNs that process two categories of representations of the input image/frame $\mathcal{I}$. Such categories consist in \textit{appearance}-based (i.e, visual) representations and a \textit{shape}-based representations. 

In both the cases, we do not consider the whole $\mathcal{I}$, but only the rectangular area that is covered by the target face. We localize the face first, and then we crop the image accordingly. This choice is crucial when processing inputs with multiple faces or when the face is not well positioned at the center of the image (or more generally, at a position incoherent with the training data). 
The \textit{appearance}-based representation of the face is simply a grayscale instance of the cropped face. In the case of the \textit{shape}-based representation, we still focus on the same cropped region, but we extract a set of shape features that essentially describe the contours of the face parts, and that, in this work, consist of a set of facial landmark points. 
However, instead of stacking their 2D coordinates into a vector (that is only possible if the set of points is consistent among different faces), we consider a more generic approach in which the shape is simply represented by an artificial image with uniform background and in which the landmarks points are depicted at their coordinates. This allows us to treat the shape in a way that is similar to what we do with the appearance, and it opens the possibility of providing different shape ``sketches'' that are not only based on landmark points (but also on contour lines, for example).

\begin{figure}
	\includegraphics[scale=0.45]{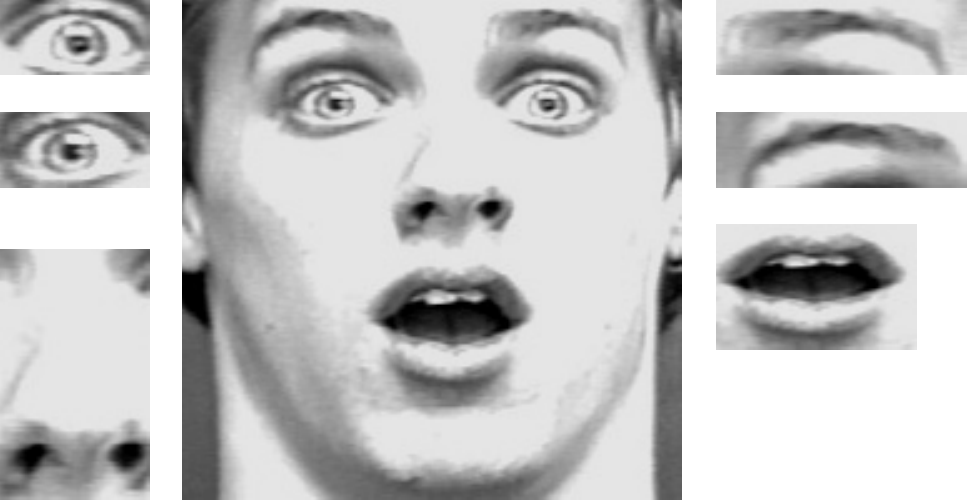}
	\hskip 1cm
	\includegraphics[scale=0.45]{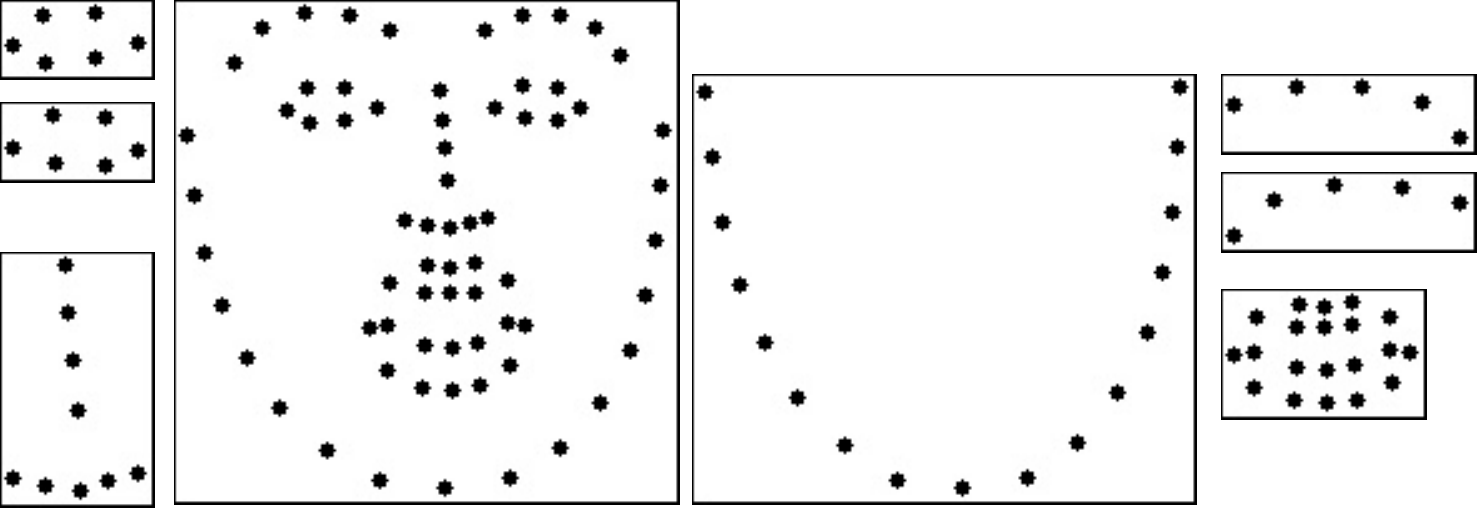}\\
	\hphantom{xxx}\textit{{appearance}}-based inputs
	\hskip 3.5cm
	\textit{shape}-based inputs
	\caption{Representations extracted from an input image. 
	On the left there are the 7 appearance-based representations. 
	On the right there are the 8 shape-based representations, that we implement by sketching landmark points in artificial images.}
	\label{sub-images}
\end{figure}

In order to study the effects of the different face parts in the recognition process, we computed the appearance and shape representations for the face (as just described) and for all the face parts: mouth, nose, eyes, eyebrows. We localized the face area and a set of 68 landmark points using the localizer of Viola and Jones \cite{viola2001rapid} and a landmark detector \cite{kazemi2014one}\footnote{We used OpenCV \url{https://opencv.org/} and the ``dlib'' library \url{http://dlib.net/}}. The detector uses the classic Histogram of Oriented Gradients (HOG) features combined with a linear classifier, an image pyramid, and a sliding window detection scheme. 
Cropping around each set of part-related landmarks (adding a small padding), we obtained $7$ instances of appearance-based representations of the input $\mathcal{I}$ and $8$ shape-based ones, since in the case of shape we also included the landmarks associated to the jaw contour. Figure \ref{sub-images} shows the overall $15$ representations that we generate. We resized these representations to the following sizes: face area $200\times200$, mouth area $80\times50$, eye area $60\times30$, eyebrow area $100\times30$, nose area $60\times100$ pixels, jaw area $200\times170$.

\vspace{-0.2cm}

\begin{figure}
	\centering 
	\includegraphics[width=0.9\textwidth]{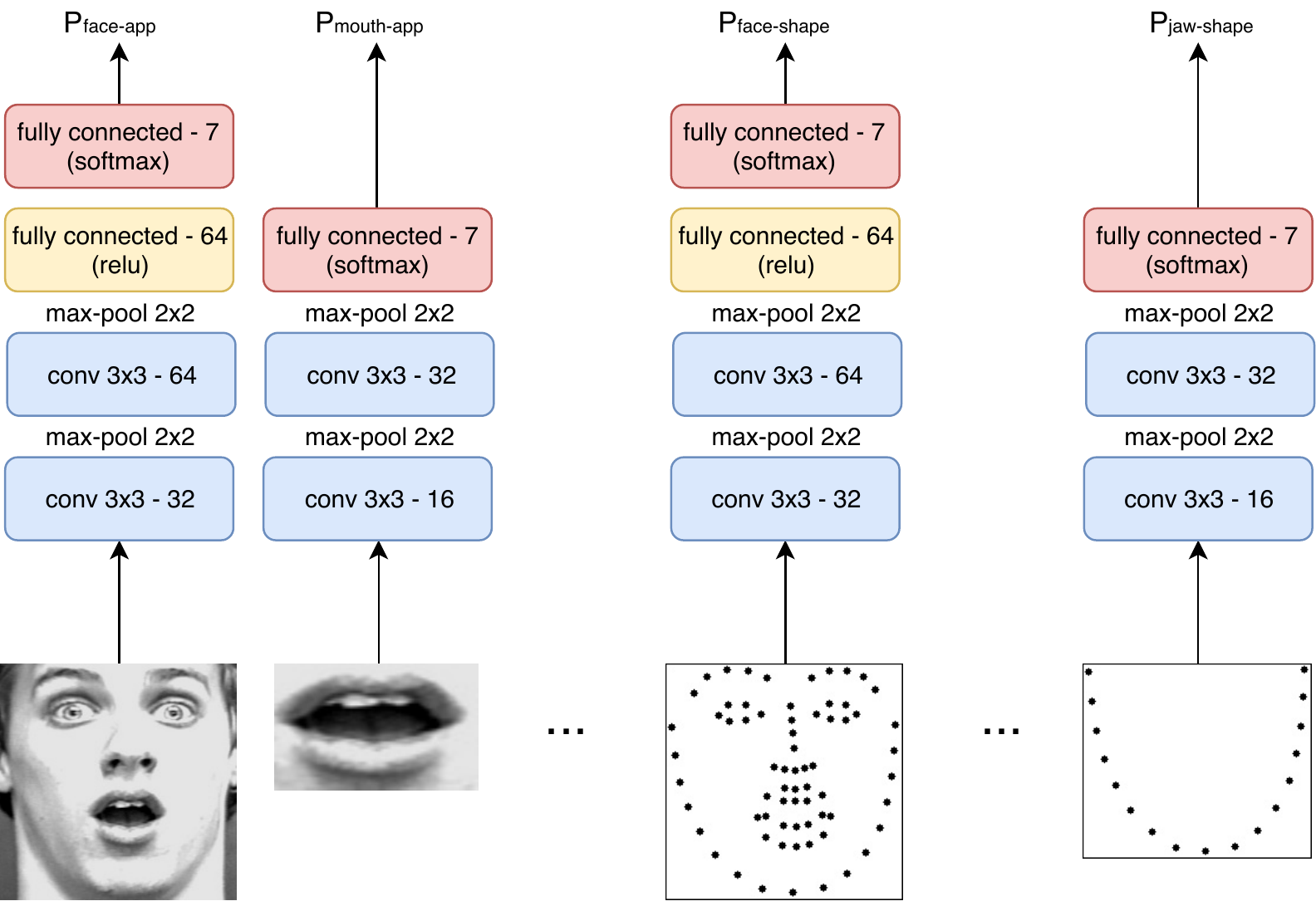}
	\caption{Structure of CNNs employed.}
	\label{CNNs}
\end{figure}

We implemented a pool of 15 CNNs, each of them processing one of the aforementioned representations (Figure \ref{CNNs}). The generic CNN$_h$ associated to the $h$-th representation has two convolutional layers followed by max pooling, and some fully connected layers terminated with a softmax activation that outputs a probability distribution over the emotions in $Y$. We indicate with $p_h(\cdot)$ the function computed by such CNN$_h$. All the hidden neural units have ReLu activation functions. The face-related CNNs have 32 and 64 filters on the two convolutional layers, respectively, and two fully connected layers (64 and $|Y|=7$ neurons). The other CNNs, that are based on inputs with smaller sizes, exploit 16 and 32 filters, and a single fully connected layer ($|Y|=7$ neurons).

The output of each of the 15 CNNs, when followed by an $\arg\max$ operation (assuming 1-based indexing), is a possible instance of the function $f$ in Eq. (\ref{a}) and Eq. (\ref{b}). Formally, for a given $h$,
\begin{eqnarray*}
x_h &=& \texttt{representation}_h(\mathcal{I})\\
p_h(x_h) &=& \texttt{CNN}_h\left(x_h\right)  \\
f(\mathcal{I}) &=& \arg\max p_h(x_h)\ ,
\end{eqnarray*}
where $x_h$ is the $h$-representation of the input, and $p_h(x_h)$ outputs a vector of size $|Y|$ that sums to $1$.
Even if our final goal is to focus on the case in which $h$ is the index of the full-face-based classifier, in Section \ref{sec:experiments} we will evaluate the quality of multiple instances of $f$, considering the predictors on the face parts too. In the next Section we will introduce a link between the full-face and face parts.


\section{Learning by Enforcing Coherence}
\label{sec:coherence}

We trained the pool of CNNs by minimizing an objective function involving the cross-entropy $L(p_h(x_{h}),y)$ between the outputs of the networks and the available labels (one-hot encoding), considering the training data $\mathcal{T} \subset \mathcal{D}$. The cross-entropy only exploits the labeled pairs in $\mathcal{T}$. However, our objective function is also composed by the penalties associated to the fulfilment of ``coherence constraints'' that we enforce on all the samples of $\mathcal{T}$, being them labeled or not.
We have considered three types of coherence, namely ``temporal coherence'', ``coherence among the predictions on the face parts'' and ``coherence between appearance and shape''.
The former enforces the CNNs to be coherent over time for each video sequence, i.e., it enforces the predictions to smoothly change along the time axis. This constraint introduces a regularizing effect, since it prevents the system from developing unstable models that abruptly change their decisions among consecutive frames\footnote{We remark that the enforcement of both the coherence constraints only happens at training time.}. The part-based coherence enforces each full-face-representation-based classifiers to take decisions that are coherent with the ones taken (\textit{on average}) by the other part-based classifiers (and vice-versa). The idea behind this constraint is that the committee of the local (i.e. part-based) predictors could provide important fine-grained information that the global (face-based) predictor might not have been able to capture. The coherence between appearance and shape enforces the prediction of the appearance-based classifier to be coherent with the prediction of the shape-based classifier for each part (excluded jaw). 

We already experimented some related constraints in the case of multi-view object recognition \cite{melacci2009semi}, and these ideas are borrowed by the generic framework of ``Learning from Constraints'' \cite{gnecco2015foundations}, where a predictor is constrained exploiting high-level knowledge on the task at hand, bridging the symbolic and sub-symbolic worlds.

In detail, given three scalars $\lambda_t,\lambda_c, \lambda_r \geq 0$ that weigh the importance of the coherence (soft) constraints, we define our objective function as the sum of the contributions (cross-entropy, temporal coherence, part coherence) of the appearance and shape representation and of the coherence between appearance and shape.
We write each contribution for the appearance-based representation (for the shape-based one it is equivalent):
\begin{small}
\begin{equation}\label{cross-entropy}
\texttt{cross-entropy}_{app} = \sum_{h}  \sum_{\substack{i=1 \\ y_i \neq \texttt{none}}} w_i \cdot L(p_{h,{app}}(x_{h,{app},i}),y_i),
\end{equation}
$$\texttt{temporal-coherence}_{app} = \sum_{h} \sum_{z=1}^{v} \sum_{t=2}^{|\mathcal{V}_z|} \left(1 - p_{h,{app}}(x_{h,{app},(z,t-1)})' \cdot p_{h,{app}}(x_{h,{app},(z,t)})\right),$$
$$\texttt{part-coherence}_{app} = \sum_{h\neq \texttt{face}} \sum_{i}\left(1 - p_{\texttt{face}, app}(x_{h,{app},i})' \cdot p_{h,{app}}(x_{h,{app},i})\right).$$
\end{small}
The index $h$ spans over the 7 appearance-based classifiers (or the 8 shape-based classifiers). The index $i$ spans over all the pairs in $\mathcal{T}$, and, for the sake of simplicity, we used the notation $y_i \neq \texttt{none}$ to indicate that we consider only the labeled examples. The scalar weights $w_i$ are used to give custom weights to the examples, and we used them to give more importance to the classes that are less represented in $\mathcal{T}$. The notation $(z,t)$ is the index of the $t$-th frame in the $z$-th video sequence belonging to $\mathcal{T}$. Finally, $\texttt{face}$ is used to indicate the index associated with the full-face input, and $'$ is the transpose operator. 

We notice that since $p_{\cdot}(\cdot)$ is a probability distribution, the dot products involving two instances of $p_{\cdot}(\cdot)$ are $1$ when such instances are equivalent (and the coherence constraints are fulfilled). 
The temporal constraint involves dot products between the predictions on pairs of consecutive frames in the same video clip.
We kept the same structure to build the part-based constraint, where the averaging operation on the part-based classifiers is evident when $\sum_{h\neq\texttt{face}}$ is moved right before the second term of the dot product $p_{\texttt{face}}(x_{h,i})' \cdot p_h(x_{h,i})$.   

Then we define the loss for the appearance-based representation (equivalently for the shape) as the sum of the three contributes defined above:
\begin{small}
\begin{equation*}
\texttt{loss}_{app} =\texttt{cross-entropy}_{app} + \lambda_t \texttt{temporal-coherence}_{app} + \lambda_c\texttt{part-coherence}_{app}.
\end{equation*}
\end{small}
Now we introduce the coherence between appearance and shape:
\begin{small}
\begin{equation}\label{loss_ag}
\texttt{coeherence}_{app-shape}= \sum_{h \neq \texttt{jaw}} \sum_i (1 - p_{h,{app}} (x_{h,{app},i})' \cdot 
p_{h,{shape}} (x_{h,{shape},i})).
\end{equation}
\end{small}
We excluded jaw, because we don't have appearance representation of it.
The final loss is the sum of all contributes just defined:
\begin{equation}\label{final_loss}
\texttt{LOSS}=\texttt{loss}_{app} + \texttt{loss}_{shape} + \lambda_r \texttt{coeherence}_{app-shape}. 
\end{equation}

\section{Experimental Results}
\label{sec:experiments}

In order to validate our model, we used the popular Extended Cohn-Kanade dataset (CK+) \cite{lucey2010extended}. It consists of 593 frames belonging to a set of short video sequences, where 120 subjects (different age and gender) generate expressions belonging to the following list: anger, contempt, disgust, fear, happiness, sadness and surprise. We excluded the sequences associated to ``contempt'', which is not included into the six universal emotions. The video sequences are composed of 10-60 frames, they start with a neutral expression and they end with the peak of one of the previously listed expressions. 
Each sequence is associated with an emotion label. 

In order to build the Semi-Supervised set $\mathcal{D}$ described in Section \ref{sec:rec}, we selected $\alpha=0.1$  and $\beta=0.7$.
We generated 5 randomizations of the whole dataset, and divided each of them into training ($70\%$), validation ($15\%$), and test sets ($15\%$), keeping the original distribution of the classes in each set. The validation data was used to validate the model parameters and excluded from training. The test partition was used to measure the quality of the model, and the results presented in this Section are averaged over the 5 test partitions (when available, we also report the standard deviation in brackets).
Each collection of training data consists of about $\approx4,000$ frames, out of which $\approx1,500$ are labeled, and they are organized into $\approx200$ sequences, while the validation data is composed of $\approx600$ frames, out of which $\approx 200$ are labeled, and organized into $\approx 30$ sequences. Since examples from the ``neutral'' class are much more represented with respect to other examples, we set $w_i=0.1$ in Eq. (\ref{cross-entropy}) if $i$ is an example from the neutral class, $w_i=1$ otherwise.
Initially we excluded coherence between appearance and shape, setting $\lambda_r=0$. We selected the optimal $\lambda_c,\lambda_t$ by a grid-search in $\left\lbrace 10^{-10}, 10^{-8}, 10^{-7}, 10^{-6}, 10^{-4}, 10^{-2}\right\rbrace$, measuring frame-level accuracy (i.e., only the labeled validation frames are considered).
We implemented our model using TensorFlow, and we minimized Eq. (\ref{final_loss}) by the Adam-based optimizer (starting learning rate $0.001$), mini-batches of size 96, and we have trained the model for multiple epochs, stopping the procedure when the validation error started increasing. 

We performed experiments comparing a system with no-coherence-constraints ($\lambda_c=\lambda_t=0$) with other models that include either temporal or part-based coherence. We compared the cases of single-frame-level predictions (where only the labeled portion of the test set is considered) and the case of video-sequence-level predictions, following the decision rules of Eq. (\ref{a}) and Eq. (\ref{b}), respectively (where $W$ covers the full video sequence).
Since examples of the different classes are not balanced in the given dataset, and in order to provide a more informative set of results, we measured two types of accuracies, namely \texttt{Micro} and \texttt{Macro} accuracies. The former is simply the percentage of correctly labeled frames/sequences, while the latter is the average of the percentages of correctly labeled frames/sequences in each emotion class. 

Table \ref{table1} shows the results we obtain when testing the classifiers that operate on the full-face inputs, considering both appearance and shape representations.
We also report results of an additional classifier obtained by averaging the outputs of the full set of $15$ classifiers (thus mixing appearance and shape data).
\begin{table}
	\caption{Micro and macro accuracies (std dev. in brackets) at image and video (sequence) level of the full-face-based classifiers (appearance and shape representations) and of an ensemble of the $15$ classifiers (average of $15$ outputs, both shape and appearance). Results without coherence constraints (\textsc{None}), with \textsc{Part}-based coherence and \textsc{Temp}-oral coherence (results where coherence improves the accuracy are in bold).}
	\centering
	{\scriptsize
	\begin{tabular}{|l|ccc|ccc|} 
	\cline{2-7}
	\multicolumn{1}{c|}{} & \multicolumn{6}{c|}{\textsc{Images}} \\
	\multicolumn{1}{c|}{} & \multicolumn{3}{c|}{\textit{\% Micro Acc}}  &\multicolumn{3}{c|}{\textit{\% Macro Acc}} \\
	\cline{2-7}
	\multicolumn{1}{c|}{} &  \textsc{None} & \textsc{Part} & \textsc{Temp} & \textsc{None}& \textsc{Part} & \textsc{Temp} \\
	\hline
	\texttt{Face}$_{app.}$ &78.9 {\tiny(3.6)} &78.0 {\tiny(2.0)} &\textbf{81.1} {\tiny(3.0)} &71.2 {\tiny(2.8)} &\textbf{72.8} {\tiny(2.2)} &\textbf{72.2} {\tiny(7.4)} \\
	\texttt{Face}$_{shape}$ &71.8 {\tiny(3.0)} &\textbf{71.9} {\tiny(3.1)} &\textbf{72.5} {\tiny(2.9)} &61.1 {\tiny(2.9)} &\textbf{61.3} {\tiny(3.0)} &\textbf{62.1} {\tiny(2.7)}\\  
	\texttt{Avg}$_{all}$ &73.7 {\tiny(4.1)} &71.4 {\tiny(3.1)} &72.1 {\tiny(4.8)} &71.9 {\tiny(3.9)} &70.2 {\tiny(3.3)} &69.7 {\tiny(3.7)} \\
	\hline
	\multicolumn{7}{c}{$\ $}\\
	[-1mm]\cline{2-7}	
	\multicolumn{1}{c|}{} & \multicolumn{6}{c|}{\textsc{Videos}} \\
	\multicolumn{1}{c|}{} & \multicolumn{3}{c|}{\textit{\% Micro Acc}}  &\multicolumn{3}{c|}{\textit{\% Macro Acc}} \\
	\cline{2-7}
	\multicolumn{1}{c|}{} &  \textsc{None} & \textsc{Part} & \textsc{Temp} &\textsc{None} & \textsc{Part} & \textsc{Temp} \\
	\hline
	\texttt{Face}$_{app.}$ &75.3 {\tiny(5.1)} &\textbf{77.0} {\tiny(3.4)} &\textbf{80.0} {\tiny(2.9)} &64.0 {\tiny(3.2)} &\textbf{66.8}{\tiny(3.1)} &\textbf{64.4} {\tiny(10.3)} \\
	\texttt{Face}$_{shape}$ &68.5 {\tiny(3.0)} &68.1 {\tiny(3.1)} &\textbf{69.4}{\tiny(2.9)} &54.0 {\tiny(2.9)} &53.5 {\tiny(3.0)} &\textbf{55.5} {\tiny(2.7)} \\ 
	\texttt{Avg}$_{all}$  &78.3 {\tiny(4.9)} &77.9 {\tiny(2.5)} &\textbf{80.4} {\tiny(5.5)} &65.6 {\tiny(6.5)} &\textbf{65.9} {\tiny(3.9)} &64.8 {\tiny(7.4)} \\ 
	\hline
	\end{tabular}}
		\label{table1}
\end{table}

Temporal coherence always improves the quality of the face-based classifiers, up to $5\%$ in the case of sequences (micro). In the case of macro-accuracy we observe larger standard deviations, that are due to the effects of the predictions on the classes with a smaller number of examples. Such classes are less-frequently predicted, and asking for a strong temporal regularization sometimes further reduces such frequency.
Coherence among parts helps in a less evident manner, especially when using shapes. Shape is less informative than appearance, resulting in a performance drop of $\approx 10\%$.
The average-based classifier is only in some cases better that the face-based ones. Constraints are less effective in this case (even if we get a strong micro accuracy in videos + temporal coherence). This suggests that mixing the $15$ classifiers together is not a promising direction, mostly because some of them have low performances that can degrade the average quality of the system. 
\begin{table}[th]
	\caption{Micro and macro accuracies (std dev. in brackets) at image and video level of all the part-based classifiers (appearance and shape representation). Results without coherence constraints (\textsc{None}), with \textsc{Part}-based coherence and \textsc{Temp}-oral coherence (results where coherence improves the accuracy are in bold).}
	\centering
	{\scriptsize
	\begin{tabular}{|l|ccc|ccc|} 
	\cline{2-7}
	\multicolumn{1}{c|}{} & \multicolumn{6}{c|}{\textsc{Images}} \\
	\multicolumn{1}{c|}{} & \multicolumn{3}{c|}{\textit{\% Micro Acc}}  &\multicolumn{3}{c|}{\textit{\% Macro Acc}} \\
	\cline{2-7}
	\multicolumn{1}{c|}{} &  \textsc{None} & \textsc{Part} & \textsc{Temp} &\textsc{None} & \textsc{Part} & \textsc{Temp} \\
	\hline
		\texttt{Mouth}$_{app.}$ &70.5 {\tiny(3.5)} &68.6 {\tiny(3.0)} &\textbf{72.8} {\tiny(2.6)} &71.5 {\tiny(6.7)} &70.8 {\tiny(5.8)} &\textbf{73.3} {\tiny(4.4)} \\ 
		\texttt{Left-eye}$_{app.}$ &42.3 {\tiny(6.0)} &41.4 {\tiny(6.0)} &40.0 {\tiny(4.2)} &41.3 {\tiny(6.5)} &39.1 {\tiny(4.9)} &38.5 {\tiny(3.9)} \\ 
		\texttt{Right-eye}$_{app.}$ &42.0 {\tiny(5.6)} &42.0 {\tiny(7.3)} &40.6 {\tiny(5.2)} &40.8 {\tiny(5.7)} &40.5 {\tiny(5.7)} &38.8 {\tiny(5.6)} \\ 
		\texttt{Left-eyebrow}$_{app.}$ &40.5 {\tiny(6.8)} &37.7 {\tiny(7.3)} &38.4 {\tiny(9.1)} &40.1 {\tiny(6.1)} &37.4 {\tiny(7.5)} &37.6 {\tiny(8.4)} \\ 
		\texttt{Right-eyebrow}$_{app.}$ &40.1 {\tiny(2.5)} &39.7 {\tiny(2.4)} &\textbf{40.4} {\tiny(2.9)} &40.1 {\tiny(3.5)} &39.5 {\tiny(2.8)} &\textbf{40.3} {\tiny(3.1)} \\ 
		\texttt{Nose}$_{app.}$ &43.6 {\tiny(2.9)} &\textbf{44.1} {\tiny(5.5)} &43.4 {\tiny(4.0)} &41.6 {\tiny(3.4)} &\textbf{42.4} {\tiny(4.8)} &\textbf{42.0} {\tiny(3.7)} \\ 
		\hline
		\texttt{Mouth}$_{shape}$ &64.3 {\tiny(2.3)} &63.8 {\tiny(3.5)} &63.4 {\tiny(3.2)} &64.4 {\tiny(4.7)} &63.4 {\tiny(4.8)} &\textbf{66.2} {\tiny(4.9)} \\ 
		\texttt{Left-eye}$_{shape}$ &35.8 {\tiny(3.4)} &34.5 {\tiny(3.7)} &35.2 {\tiny(2.6)} &33.2 {\tiny(3.9)} &33.0 {\tiny(3.4)} &32.5 {\tiny(2.3)} \\ 
		\texttt{Right-eye}$_{shape}$ &40.7 {\tiny(3.2)} &40.6 {\tiny(2.7)} &\textbf{41.5} {\tiny(3.0)} &36.9 {\tiny(2.4)} &\textbf{37.2} {\tiny(2.1)} &\textbf{37.9} {\tiny(2.0)} \\ 
		\texttt{Left-eyebrow}$_{shape}$ &31.2 {\tiny(4.4)} &31.0 {\tiny(3.8)} &30.1 {\tiny(3.5)} &31.8 {\tiny(1.8)} &\textbf{31.9} {\tiny(2.0)} &31.7 {\tiny(3.7)} \\ 
		\texttt{Right-eyebrow}$_{shape}$ &34.3 {\tiny(4.2)} &33.9 {\tiny(3.7)} &34.1 {\tiny(3.5)} &34.3 {\tiny(5.2)} &33.4 {\tiny(4.5)} &33.6 {\tiny(4.9)} \\ 
		\texttt{Nose}$_{shape}$ &30.8 {\tiny(3.7)} &30.4 {\tiny(3.2)} &\textbf{30.9} {\tiny(4.2)} &30.6 {\tiny(5.6)} &\textbf{31.0} {\tiny(5.0)} &\textbf{31.6} {\tiny(5.2)} \\ 
		\texttt{Jaw}$_{shape}$ &37.4 {\tiny(3.7)} &37.2 {\tiny(3.7)} &37.0 {\tiny(3.5)} &34.1 {\tiny(4.6)} &\textbf{34.9} {\tiny(4.3)} &33.8 {\tiny(4.0)}\\ 
	\hline
	\multicolumn{7}{c}{$\ $}\\
	[-1mm]\cline{2-7}
	\multicolumn{1}{c|}{} & \multicolumn{6}{c|}{\textsc{Videos}} \\
	\multicolumn{1}{c|}{} & \multicolumn{3}{c|}{\textit{\% Micro Acc}}  &\multicolumn{3}{c|}{\textit{\% Macro Acc}} \\
	\cline{2-7}
	\multicolumn{1}{c|}{} &\textsc{None}   & \textsc{Part} & \textsc{Temp} & \textsc{None}& \textsc{Part} & \textsc{Temp} \\	
	\hline
		\texttt{Mouth}$_{app.}$ &77.5 {\tiny(7.7)} &72.3 {\tiny(9.0)} &75.7 {\tiny(6.4)} &73.0 {\tiny(9.5)} &66.4 {\tiny(8.4)} &69.9 {\tiny(8.7)} \\ 
		\texttt{Left-eye}$_{app.}$  &49.4 {\tiny(8.4)} &\textbf{50.6} {\tiny(4.1)} &47.2 {\tiny(5.9)} &42.7 {\tiny(5.8)} &41.3 {\tiny(2.7)} &40.2 {\tiny(6.3)} \\ 
		\texttt{Right-eye}$_{app.}$ &46.8 {\tiny(2.3)} &\textbf{47.2} {\tiny(4.9)} &\textbf{47.7} {\tiny(2.9)} &39.8 {\tiny(1.7)} &39.2 {\tiny(3.0)} &38.9 {\tiny(3.7)} \\ 
		\texttt{Left-eyebrow}$_{app.}$ &43.0 {\tiny(9.7)} &41.7 {\tiny(9.2)} &42.1 {\tiny(11.1)} &35.2 {\tiny(7.7)} &34.3 {\tiny(9.1)} &34.3 {\tiny(9.6)} \\ 
		\texttt{Right-eyebrow}$_{app.}$ &43.4 {\tiny(4.6)} &42.5 {\tiny(5.5)} &\textbf{43.8} {\tiny(3.2)} &36.5 {\tiny(6.6)} &35.6 {\tiny(6.8)} &35.9 {\tiny(4.0)} \\ 
		\texttt{Nose}$_{app.}$ &44.3 {\tiny(4.9)} &\textbf{47.7} {\tiny(5.1)} &\textbf{47.2} {\tiny(2.8)} &35.4 {\tiny(4.3)} &\textbf{38.8} {\tiny(4.3)} &\textbf{38.9} {\tiny(3.1)} \\ 
		\hline
		\texttt{Mouth}$_{shape}$ &71.9 {\tiny(2.5)} &\textbf{74.0} {\tiny(3.7)} &70.6 {\tiny(2.8)} &64.3 {\tiny(4.2)} &\textbf{66.1} {\tiny(6.0)} &\textbf{67.3} {\tiny(5.0)} \\ 
		\texttt{Left-eye}$_{shape}$ &45.1 {\tiny(5.8)} &44.7 {\tiny(8.5)} &45.1 {\tiny(4.5)} &36.6 {\tiny(7.1)} &\textbf{37.2} {\tiny(6.1)} &\textbf{38.3} {\tiny(4.1)} \\ 
		\texttt{Right-eye}$_{shape}$ &51.9 {\tiny(2.2)} &\textbf{52.8} {\tiny(3.7)} &\textbf{56.2} {\tiny(3.7)} &39.4 {\tiny(3.1)} &\textbf{41.5} {\tiny(3.3)} &\textbf{44.9} {\tiny(3.9)} \\ 
		\texttt{Left-eyebrow}$_{shape}$ &36.2 {\tiny(6.7)} &34.5 {\tiny(3.4)} &34.9 {\tiny(3.5)} &28.7 {\tiny(5.1)} &28.7 {\tiny(3.0)} &\textbf{29.3} {\tiny(4.1)} \\ 
		\texttt{Right-eyebrow}$_{shape}$ &40.4 {\tiny(5.0)} &40.0 {\tiny(5.9)} &\textbf{41.3} {\tiny(6.7)} &33.9 {\tiny(5.6)} &33.1 {\tiny(5.0)} &33.8 {\tiny(7.0)} \\ 
		\texttt{Nose}$_{shape}$ &37.5 {\tiny(5.0)} &35.7 {\tiny(3.7)} &34.0 {\tiny(1.4)} &31.4 {\tiny(5.4)} &28.5 {\tiny(5.6)} &\textbf{31.8}{\tiny(4.4)} \\ 
		\texttt{Jaw}$_{shape}$ &40.9 {\tiny(2.5)} &40.9 {\tiny(2.1)} &40.0 {\tiny(3.7)} &30.5 {\tiny(2.5)} &\textbf{31.3} {\tiny(2.7)} &29.8 {\tiny(2.7)} \\ 
		\hline
	\end{tabular}}
		\label{table2}
\end{table}

To gain better insights about the last comment, Table \ref{table2} reports the accuracies for all the part-based classifiers. 
The mouth area is a very effective input for facial expression recognition, that can sometimes compete with the full-face. 
This is more evident in the case of videos, when comparing shape-based representations of face and mouth.
As expected, the other parts are worse than the full-face, since they are just local views.
The addition of both coherences sparsely helps in improving the local classifiers, with a preference toward temporal coherence.
The worst results are obtained by eyebrows and nose in shape-based classification.
Interestingly, the eye-based predictors score the most effective results after face and mouth in video sequences. While their appearance representation is altered when eyes get closed, their shape representation is more stable. 
The results on left eye and right eye are a bit different and this due to the fact that wrinkles can be asymmetric, or that an eye can be closed, or to the variation of lighting and pose.
This analysis suggests that an accurate choice of a sub-portion of the face parts could significantly help the part-based coherence constraint (since some of the parts are not very informative).

\begin{figure}
	\centering
	\includegraphics[width=	10cm]{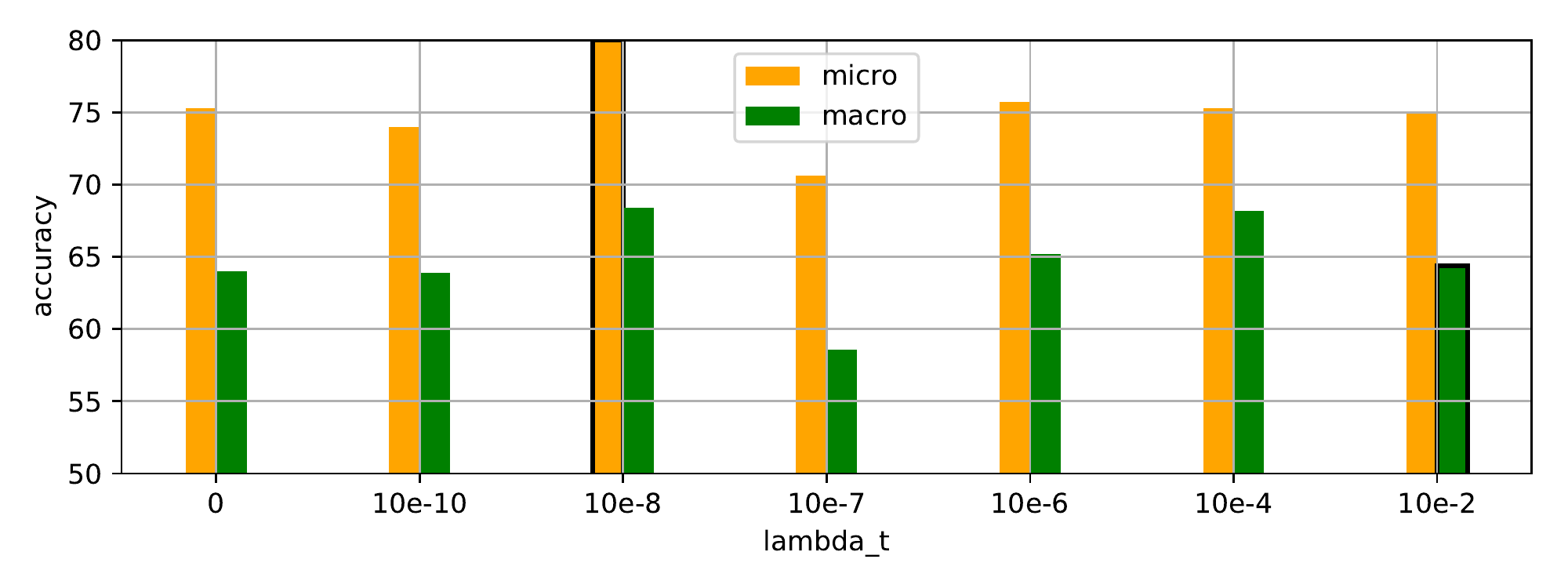}
	\caption{Micro and macro accuracies in the case of video data, full-face-based classifier (appearance), for different values of $\lambda_t$. The black-bordered bars are the results we reported in Table \ref{table1}.}
	\label{graphic}
\end{figure}

We deepened the analysis on the temporal-constrained classifiers in the case of making predictions in video sequences.
Since the number of sequences is small, we selected the optimal $\lambda_t$ using image-level predictions on the validation data (as already stated), leading to $\lambda_t=10^{-8}$ and $\lambda_t=10^{-2}$ in the case of micro and macro accuracy, respectively. Figure \ref{graphic} reports the performances on videos with different values of $\lambda_t$ (appearance only).
We can see that the distributions of the performances are multimodal, and if we focus on the macro accuracy we observe that we could have obtained much better results with different values of $\lambda_t$. This suggests that the validation procedure has room for being improved in the case of video data.

\begin{figure}
	\centering
	\includegraphics[width=1\textwidth]{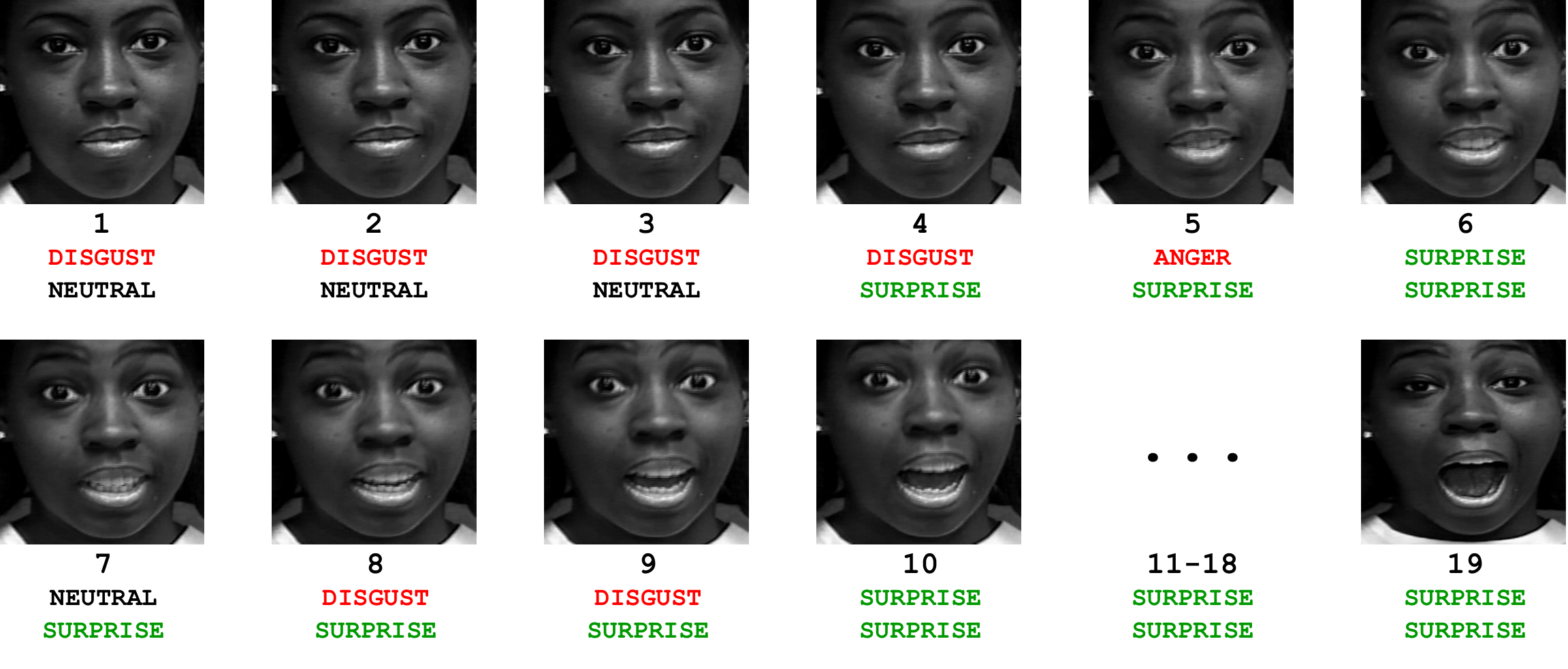}
	\caption{Predictions in a sequence that starts with neutral expression and develops in surprise. For each frame we report the prediction of the model without coherences (top) and the prediction with temporal coherence (bottom). The wrong predictions are in red.}
	\label{esempio_qualitativo}
\end{figure}

Temporal coherence yields homogeneous predictions over the sequences, without oscillations along the temporal axis.   
In Figure \ref{esempio_qualitativo} we represent an example showing that temporal coherence produces an uniform trend in the predictions on the sequence (``surprise'' emotion is sketched). In fact the model with the best $\lambda_t$ predicts ``neutral'' in the first frames of the sequence and ``surprise'' in the last ones. Differently, the model without temporal coherence produces an oscillating trend on the sequence, predicting also wrong emotions as ``disgust'' and ``anger''.

In Table \ref{table3} we show the results on single emotion classes for face and mouth appearance-based classification, focussing on the case where no-coherence is introduced and the ones with a selection of the best $\lambda_c>0$ and $\lambda_t>0$ from the previously described experiments. 
``Fear'' and ``sadness'' classes are difficult to classify because they do not involve strong facial movements, while ``happiness'' and ``surprise'' are easy to recognize. 
The mouth-based model has difficulties in the ``neutral'' class, since some emotions do not evidently alter the mouth area (the face model does not show this issue). In the ``sadness'' class, where the face-based model scores low accuracies, the mouth-based classifier is much more performant. 
This suggests that the face-related network has difficulties in developing a generalizable representation for the whole face to identify ``sadness''. Larger training data could help in this case. 
\begin{table} 
	\caption{Accuracies on each class of full-face and mouth classifiers (appearance).
		Results without coherence constraints, with \textsc{Part}-based coherence and \textsc{Temporal} coherence (results where coherence improves the accuracy are in bold).}
	\centering
	{\scriptsize
	\begin{tabular}{|l|ccccccc|}
		\cline{2-8}
		\multicolumn{1}{c|}{} & \multicolumn{7}{|c|}{\textsc{Images}} \\
		\multicolumn{1}{c|}{}  & \textit{Anger}	& \textit{Disgust} &	\textit{Fear} &	\textit{Happiness} &	\textit{Sadness} &	\textit{Surprise} &	\textit{Neutral}\\
		\hline
		\texttt{face}$_{app.}$ \textsc{None} &73.7 &	69.2 &	56.1 &	92.5 &	29.5 &	96.1 &	81.1 \\
		\texttt{face}$_{app.}$ \textsc{Part} &68.0 &	\textbf{78.2} &	\textbf{75.2} &	\textbf{98.2} &	24.3 &	\textbf{97.4} &	68.6 \\
		\texttt{face}$_{app.}$ \textsc{Temp} &\textbf{77.1} &	\textbf{81.8} &	50.0 &	\textbf{97.5} &	26.2 &	95.5 &	\textbf{81.8} \\
		\hline
		\texttt{mouth}$_{app.}$ \textsc{None}$\ $ &66.4 &	69.6 &	59.4 &	92.7 &	75.6 &	96.6 &	40.2 \\
		\texttt{mouth}$_{app.}$ \textsc{Part}  &66.4 &	\textbf{81.8} &	\textbf{65.1} &	\textbf{95.0} &	59.6 &	95.5 &	32.2 \\
		\texttt{mouth}$_{app.}$ \textsc{Temp} &\textbf{67.8} &	\textbf{80.4}&	58.8 &	\textbf{94.8} &	72.0 &	95.2 &	\textbf{44.1} \\	
		\hline
	\multicolumn{7}{c}{$\ $}\\
	[-1mm]\cline{2-8}
		\multicolumn{1}{c|}{} & \multicolumn{7}{|c|}{\textsc{Videos}} \\
		\multicolumn{1}{c|}{}  & \textit{Anger}	& \textit{Disgust} &	\textit{Fear} &	\textit{Happiness} &	\textit{Sadness} &	\textit{Surprise} &	\textit{Neutral}\\	
		\hline
		\texttt{face}$_{app.}$ \textsc{None}& 77.1&	62.2 &	33.3 &	90.9 &	25.0 &	95.4& -- \\
		\texttt{face}$_{app.}$ \textsc{Part}& 68.6 &	\textbf{71.1} &	\textbf{53.3} &	90.9 &	20.0 &	\textbf{96.9} & --\\
		\texttt{face}$_{app.}$ \textsc{Temp}&77.1&	\textbf{73.3} &	\textbf{40.0}&	\textbf{98.2}&	25.0 &	\textbf{96.9} & --\\
		\hline
		\texttt{mouth}$_{app.}$ \textsc{None}$\ $& 77.1 &	73.3 &	46.7 &	78.2 &	75.0 &	87.7 & --\\
		\texttt{mouth}$_{app.}$ \textsc{Part}  & 62.9 &	\textbf{77.8} &	46.7 &	74.6 &	55.0 &	81.5 & --\\
		\texttt{mouth}$_{app.}$ \textsc{Temp} &74.3 &	\textbf{77.8} &	40.0 &	74.6 &	65.0 &	87.7 & --\\
		\hline
	\end{tabular}}
	\label{table3}	
\end{table}

Temporal coherence shows better performance in ``neutral'', ``anger'' (image-level only), and ``disgust'' emotions. It is also helpful in the ``happiness'' class, where the face model performs a close-to-flawless classification.
Introducing coherence among parts improves the recognition of ``disgust'', ``fear'' (face only), ``happiness'' (image-level only), and it slightly improves the accuracy of ``surprise'' for the face-based predictor.  

In addiction to these results, we report that eye-based recognition reaches very good results for the ``surprise'' class; the accuracy of right-eye classifier with temporal coherence is $88.2\%$.  This is due to the fact that the eyes in surprise expressions are wide open, so easily recognizable.
Differently, the ``neutral'' class is not recognizable at all from the eyebrows. 
Nose-based classification (appearance) reaches an accuracy of 79.4\% with temporal coherence in the ``disgust'' class, where the nose is wrinkled.

At a later stage, considering the previously reported experiments, we made other experiments with the coherence between appearance and shape (\ref{loss_ag}), changing some values of $\lambda_r$. We obtained that the new coherence helps further the accuracies, with respect to the best model with only temporal coherence. As we can see in Table \ref{table_ag} the model with the best $\lambda_t$ and $\lambda_r$ ($\lambda_t=\lambda_r=10^{-8}$) for full-face classification on appearance-based representation is sometimes better than the model with the best $\lambda_t$ only. Coherence between appearance and shape improves the micro and macro accuracy and the classification of some emotions as ``anger'', ``disgust'' (even 6.7\%), ``fear'' and ``surprise'' at sequence level. At frame level it improves the accuracy in ``neutral'' class even of 4\%.
\begin{table} 
	\caption{Micro and macro accuracies and accuracies on each class of full-face classifier (appearance).
		Results without coherence constraints, with \textsc{Temporal} coherence only and  with \textsc{Temporal} and coherence between appearance and shape (results where coherence between appearance and shape improves the accuracy respect to temporal coherence only are in bold).}
	\centering
	\resizebox{\linewidth}{!}{
		\begin{tabular}{|l|ccccccccc|}
			\cline{2-10}
			\multicolumn{1}{c|}{} & \multicolumn{9}{|c|}{\textsc{Images}} \\
			\multicolumn{1}{c|}{}  &  \textit{Micro}	& \textit{Macro} & \textit{Anger}	& \textit{Disgust} &	\textit{Fear} &	\textit{Happiness} &	\textit{Sadness} &	\textit{Surprise} &	\textit{Neutral}\\
			\hline
			\texttt{face}$_{app.}$ \textsc{None} & 78.9 & 71.2 &73.7  &	69.2 &	56.1 &	92.5&	29.5 &	96.1&	81.1 \\
			\texttt{face}$_{app.}$ \textsc{Temp} & 81.1 & 72.9 & 77.1 &	81.8 &	50.0&	97.5 &	26.2 &	95.5 &	81.9 \\
			\texttt{face}$_{app.}$ \textsc{Temp+app-shape} & 80.7 &	\textbf{72.9}&	73.7 &	81.2 &	49.0&	94.8& \textbf{31.3} & 94.6 &\textbf{85.8}  \\
			\hline
			\multicolumn{9}{c}{$\ $}\\
			[-1mm]\cline{2-10}
			\multicolumn{1}{c|}{} & \multicolumn{9}{|c|}{\textsc{Videos}} \\
			\multicolumn{1}{c|}{}  & \textit{Micro}	& \textit{Macro} &\textit{Anger}	& \textit{Disgust} &	\textit{Fear} &	\textit{Happiness} &	\textit{Sadness} &	\textit{Surprise} &	\textit{Neutral}\\	
			\hline
			\texttt{face}$_{app.}$ \textsc{None}& 75.3 & 64.0 & 77.1 &62.2 &	33.3 &	90.9&	25.0 &	95.4 & -- \\
			\texttt{face}$_{app.}$ \textsc{Temp}& 80.0 & 68.4 &77.1 &  73.3&	40.0 &	98.2 &	25.0 &	96.9 & --\\
			\texttt{face}$_{app.}$ \textsc{Temp+app-shape}& \textbf{80.4}&\textbf{69.9} & \textbf{80.0} &\textbf{80.0}&	\textbf{46.7}&	89.1&25.0 &\textbf{98.5}& --\\
			\hline
	\end{tabular}}
	\label{table_ag}	
\end{table}

As a final comment, we have also tried to perform some experiments involving both temporal and part-based coherences activated, and others involving the three coherences together, but they were not better than the ``best'' ones that we obtained by activating temporal coherence and coherence between appearance and shape.

\section{Occlusions}
\label{sec:occlusions}

Shape-based representation can help to recognize emotions when there are occlusions or different illuminations on the face. We did some tests in images with occlusions: we took the last frame (the more expressive) of each sequence of the CK+ dataset, so we obtained 309 images, and we covered some parts of the face, as mouth or nose. We made predictions on this modified images (appearance and shape-based representations) and we found that sometimes the shape-based predictor on face is better than the appearance one. 
In Table \ref{occlusion_table} we report the accuracies associated with the cases in which the shape-based representation performs better than appearance (full-face classifier). For each emotion and for each part the accuracies are the percentage of the right predictions on the frames with the part covered. For ``anger'', when the mouth is covered, the accuracy for the appearance-based classifier is only 35.6\% while for the shape-based one is 75.6\%. As we have seen in Section \ref{sec:experiments}, this emotion is not easy to recognize, and covering an important part as the mouth makes the task more difficult. Differently, the shape-based classifier can capture more robust features that go beyond the appearance even when the mouth is occluded. 
An other considerable improvement of the shape respect to appearance happens when nose is covered in images with happy expressions. The appearance-based classifier sometimes is confused with ``fear'' where the mouth is in general open as in ``happiness''.
\begin{table}
	\caption{Accuracies of full-face classifier (appearance and shape) on images with occlusions. }
	\centering
	\begin{tabular}{|c|c|cc|}
		\hline
		emotion & covered part & acc.$_{app}$ & acc.$_{shape}$ \\
		\hline
		\textit{anger} & 	\texttt{mouth} & 35.6 & 75.6 \\
		\textit{disgust} & 	\texttt{mouth} & 61.0 & 78.0 \\
		\textit{disgust} & 	\texttt{nose} & 81.4 & 83.1 \\
		\textit{happiness} & 	\texttt{nose} & 66.7 & 98.6 \\
		\textit{sadness} & 	\texttt{mouth} & 67.9 & 71.4 \\
		\textit{sadness} & 	\texttt{nose} & 53.6 & 75.0 \\
		\textit{surprise} & 	\texttt{nose} & 95.2 & 98.8\\
		\hline
	\end{tabular}
\label{occlusion_table}
\end{table} 

\begin{figure}
	\includegraphics[width=\textwidth]{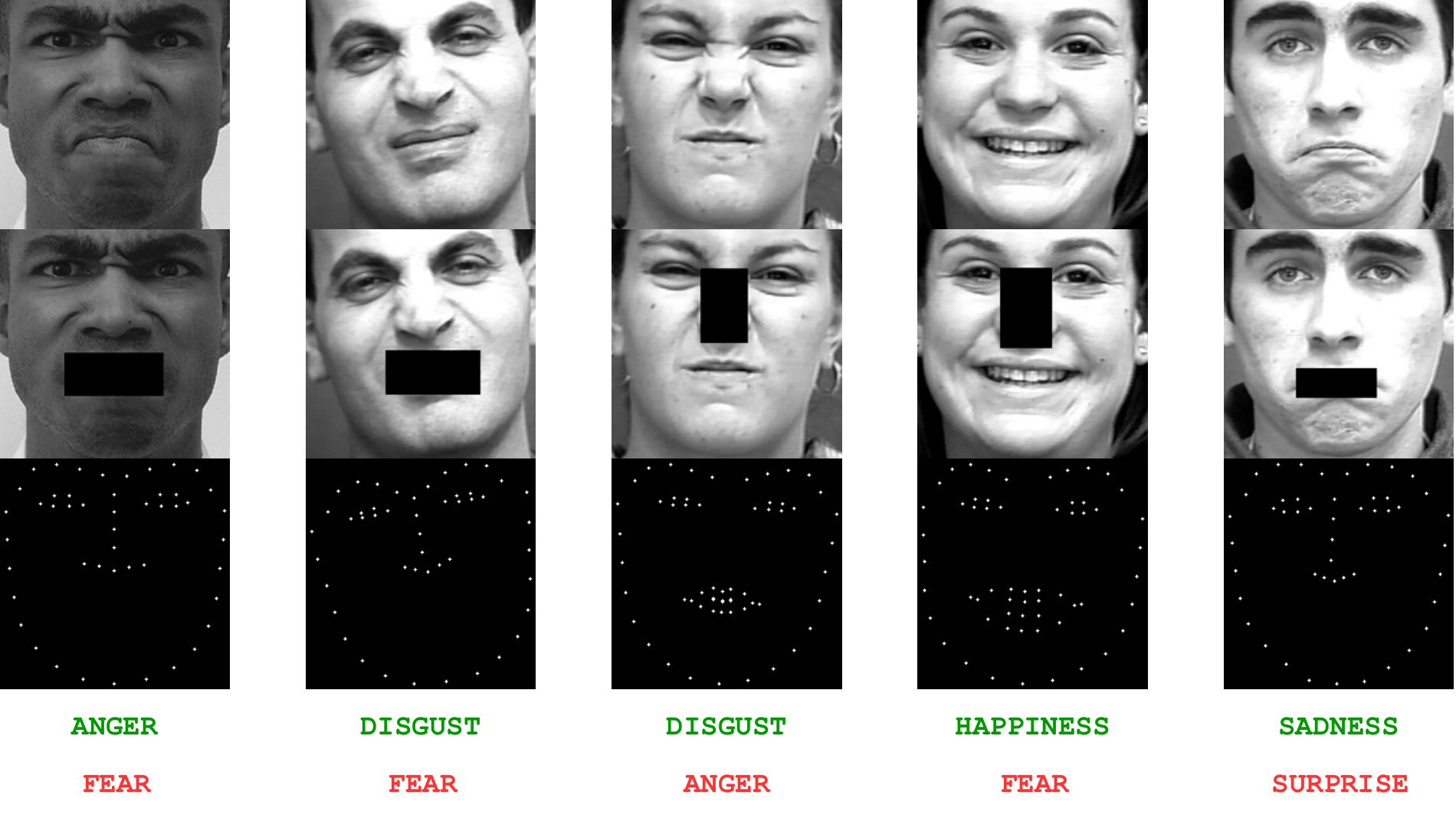}
	\caption{Examples of images with occlusions where the shape-based classifier predicts the right emotion whereas the appearance-based classifier gets wrong. From top to bottom: the original image (appearance), the images with occlusion (appearance, shape), the right prediction of the shape-based classifier (green), and the wrong prediction of the appearance-based classifier (red).}\label{occlusion_img}
\end{figure}

In Figure \ref{occlusion_img} we report some examples with occlusions in which the shape-based classifier predicts the right emotion, while the appearance-based gets wrong. 
The first example from the left represents ``anger'', whereas the appearance-based classifier predicts  ``fear'' when the mouth is covered. Anger and fear show most of their differences in the mouth area. In angry expression the lips are tight, while in fear the mouth is slightly open. 
In the second example representing  ``disgust'', where the wrinkled mouth is covered, the appearance classifier predicts ``fear''. 
In the third instance the occlusion is on the wrinkled nose typical of the disgusted expression.
In the following, covering the nose, the appearance-based classifier predicts ``fear'' instead of ``happiness'', because it focuses on the slightly open mouth, not considering the more relaxed nose. 
In the last example depicting ``sadness'' with mouth covered the appearance classifier, which focuses on the open eyes and on the slightly raised eyebrows, predicts ``surprise'' not seeing if the mouth is wide open or down.

\section{Conclusions and Future Work}
\label{sec:conclusions}

We presented a Convolutional Neural Network (CNN)-based approach to Facial Expression Recognition. Our model is based on a pool of CNNs that process distinct face parts, represented using visual (appearance) or shape-only features. In the latter case, we treated shape as a generic input of the learnable model, without manually engineering its representation.
Shape-based representation can help to detect the right expression when the appearance-based representation fails, for example in presence of occlusions on the face or of different illumination levels.

 We studied the importance of the different representations on the task at hand, showing an analysis that involved all the considered face parts, and reporting results of experiments on a popular dataset composed of six basic emotions, plus the neutral case. We proposed the introduction of coherence constraints among the face-part predictors, between predictions on consecutive time instants, and between appearance and shape representation, casting the learning problem in the Semi-Supervised setting and using video data. Our results have shown that using unsupervised training data paired with coherence constraints improves the quality of the recognizer, especially in the case of temporal coherence combining with coherence between appearance and shape. Our future work will include a more detailed study on the face-part coherence, selecting only on the most promising face parts, according to the results of this study.
 We will use a larger collections of data, to grasp the importance of large-scale unsupervised data obtained from video conferences.

\bibliographystyle{splncs03}
\bibliography{arxiv}

\end{document}